\documentclass[letterpaper, 10 pt, conference]{ieeeconf}
\usepackage{cite}
\usepackage{amsmath,amssymb,amsfonts,enumerate,mathabx}
\usepackage{algorithmic}
\usepackage{graphicx}
\usepackage{textcomp}
\usepackage{arydshln}
\usepackage{comment}
\usepackage{svg}
\def\BibTeX{{\rm B\kern-.05em{\sc i\kern-.025em b}\kern-.08em
    T\kern-.1667em\lower.7ex\hbox{E}\kern-.125emX}}

\newtheorem{prop}{Proposition}
\newtheorem{thm}{Theorem}

\newtheorem{lem}{Lemma}

\newtheorem{defn}{Definition}

\begin{document}

\IEEEoverridecommandlockouts   
\overrideIEEEmargins
\title{Nonconvex Obstacle Avoidance using \\Efficient Sampling-Based Distance Functions}

\author{Paul Lutkus, Michelle S. Chong, and Lars Lindemann 
\thanks{P. Lutkus is with the Thomas Lord Department of Computer Science, University of Southern California. {Email: \tt\small lutkus@usc.edu}}
\thanks{M. Chong is with the Department of Mechanical Engineering, Eindhoven University of Technology (TU/e).  {Email: \tt\small m.s.t.chong@tue.nl}}
\thanks{L. Lindemann is with the the Thomas Lord Department of Computer Science and the Ming Hsieh Department of Electrical and Computer Engineering, University of Southern California. {Email: \tt\small llindema@usc.edu}}
\thanks{M. Chong and L. Lindemann are thankful to the Eindhoven Artificial Intelligence Systems Institute (EAISI) for supporting L. Lindemann's visit to TU/e which sparked the development of this work. L. Lindemann was partially supported by the National Science Foundation through IIS-SLES-2417075 and Toyota R\&D through the USC Center for Autonomy and AI. }
}

\maketitle
\thispagestyle{empty} 

\begin{abstract}
We consider nonconvex obstacle avoidance  where a robot described by nonlinear dynamics and a nonconvex shape has to avoid nonconvex obstacles. Obstacle avoidance is a fundamental problem in robotics and well studied in control. However, existing solutions are computationally expensive (e.g., model predictive controllers),  neglect nonlinear dynamics (e.g., graph-based planners), use diffeomorphic transformations into convex domains (e.g., for star shapes), or are conservative due to convex overapproximations. The key challenge here is that the computation of the distance between the shapes of the robot and the obstacles is a nonconvex problem. We propose efficient computation of this distance via sampling-based distance functions. We  quantify the sampling error and show that, for certain systems, such sampling-based distance functions are valid nonsmooth control barrier functions. We also study how to deal with disturbances on the robot dynamics in our setting. Finally, we illustrate our method on a robot navigation task involving an omnidirectional robot and nonconvex obstacles. We also  analyze  performance and  computational efficiency of our controller as a function of the number of samples. 
\end{abstract}


\section{Introduction} \label{sec:introduction}



\subsection{Motivation}
 Collision avoidance in control and motion planning problems is achieved by enforcing a minimum distance between the robots and the obstacles. This involves computing the minimum distance, which can be defined as an optimization problem.  This can be solved in various ways, including approximations via signed distance fields \cite{li2024representing}, the Expanding Polytope algorithm \cite{van2001proximity} for polytopic representations or the Gilbert, Johnson and Keerthi (GJK) algorithm \cite{gilbert1988fast, cameron1997enhancing, rassaerts2024obstacle} for convex set representations of the robots and the obstacles. When the robot(s) and/or obstacles can be approximated by simple enough shapes, the minimum distance 
is a  differentiable function. For example, point-masses \cite{chen2017obstacle} and ellipsoids \cite{yan2015closed, verginis2019closed}.  For robotic systems with a complex body, approximating its body with simple shapes for computation is unable to yield the proximity to obstacles often desired in applications. This necessitates representation of the robots and obstacles with possibly non-convex shapes, which yield \textit{non-differentiable} minimum distance functions.

A popular methodology in enforcing a minimum distance is to encode the requirement in a so called safe set, and to enforce its forward invariance using Control Barrier Functions (CBF)  \cite{ames2016control}. This entails choosing the minimum distance function as the candidate CBF where collision avoidance is enforced when the gradient of the CBF along the solutions to the system satisfy the CBF conditions. We are motivated by the collision-free movement of robots and obstacles in close proximity, which requires nonconvex set representations of the robots and obstacles. This results in minimum distance functions, and consequently CBF, which are nondifferentiable. Therefore, nonsmooth analysis tools are needed and the framework from \cite{glotfelter2017nonsmooth} for non-smooth CBF will be crucial for solving the problem in this paper.  

\begin{figure}
    \centering
    \includegraphics[width=0.95\linewidth]{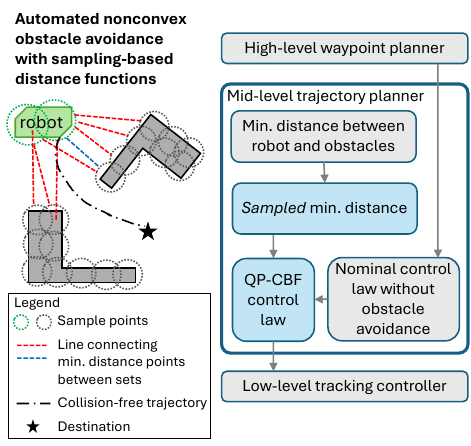}
    \caption{\textbf{(Left)} Sampling-based minimum distance functions. We sample the sets covering the robot and obstacles and compute the minimum distance between the sample points. The red and blue dashed lines connect the closest sample points between each set. The blue dashed line is the shortest distance between the robot and the obstacle. \textbf{(Right)} Our contributions are (i) sampling to approximate the minimum distance between robot and obstacles, (ii) sampling error quantification, and (iii) guaranteed obstacle avoidance with CBF-based control law using its QP formulation.   }
    \label{fig:overview}
\end{figure}

\subsection{Related work}
Robot navigation with obstacle avoidance typically deploys layered control architectures (see Figure \ref{fig:overview}) consisting of a high-level waypoint planner, a mid-level trajectory planner, and a low-level tracking controller. The high-level waypoint planner generates waypoints between  start and end locations, such that the mid-level trajectory planner can produce the control input (e.g., velocity or acceleration) to follow the planned waypoints. The high and mid-layers hence  function  together as the motion planner, which computes the state trajectory and corresponding control input that connects the start and end locations, while satisfying system dynamics and safety requirements. The low-level controller executes the control input generated by the motion planner.

Popular high-level planners \cite{lavalle2006planning} include sampling-based planners such as RRT and the Probabilistic Roadmap; and mid-level trajectory planner include trajectory optimizers (see \cite{schulman2014motion} for a review), reachability-based methods \cite{kwon2024conformalized, michaux2024safe}, and model predictive control.  Although significant advances have been made, all the aforementioned methods are computationally intensive, particularly when nonconvex obstacles are involved even rendering some approaches infeasible.

More efficient  mid-level approaches to robot navigation around obstacles are based on hybrid feedback control laws that switch from \textit{move-to-target} to \textit{obstacle-avoidance} mode when in close proximity with the obstacles, see \cite{sawant2024hybrid,berkane2021obstacle} and references therein. Other methods include 
navigation functions that instead involve constructing diffeomorphic transformations, e.g., for star-shaped sets \cite{rimon1991construction,vasilopoulos2018reactive}. While these approaches ensure deadlock-free, safe robot navigation, they typically do not generalize to arbitrary nonconvex sets.

In recent years, a popular approach in designing the mid-level trajectory planner has been the use of CBFs and  its quadratic program formulation as a \textit{safety filter} \cite{ames2016control, hobbs2023runtime, nguyen2021robust, singletary2021safety}. The CBF-based controller generates the control input to avoid collision with obstacles which were unaccounted for by the high-level planner. One notable result considers convex set representations of the robot and obstacle \cite{thirugnanam2023nonsmooth}, where the authors showed that the minimum distance function for a class of state-dependent convex sets is continuously differentiable. Hence, the standard CBF results developed in \cite{ames2016control} can be used to guarantee obstacle avoidance. 

To the best of our knowledge, no works address general nonconvex set representation of robot(s) and obstacle(s) with provable guarantees of collision-free navigation for robots with control affine system dynamics. Even under nominal conditions, the main challenge stems from the fact that the minimum distance function between nonconvex sets are hard to compute and is nondifferentiable.

\subsection{Contributions}
In our paper, we instead propose sampling the geometric set representation to obtain an approximation of the true distance function. The sampling-induced approximation error as well as the desired minimum distance are then encoded in our candidate CBF. The resulting function is non-differentiable and we use the results in \cite{glotfelter2017nonsmooth} to show that the resulting function is a valid non-smooth CBF. This generates a set of controllers, which guarantees obstacle avoidance to a desired safety margin if the controller is continuous. An illustration of our sampling-based approach in approximating the true minimum distance function and our contributions in a typical robot navigation architecture is provided in Figure \ref{fig:overview}. To bring our results closer to practical implementation, we also provide the safety guarantees in the presence of disturbances, by tightening the CBF conditions under non-restrictive assumptions on the disturbances.

The paper is structured as follows. We present  preliminaries in Section \ref{sec:prelim} where we introduce our notations and recall non-smooth CBF theory from \cite{glotfelter2017nonsmooth}. Section \ref{sec:mot-prob} motivates and states our problem formulation. Our main results are developed in Sections \ref{sec:sampling-ncbf} and \ref{sec:disturbance-errors}, and we illustrate them in simulations in Section \ref{sec:sim}. Section \ref{sec:conclude} concludes our paper.

\section{Preliminaries} \label{sec:prelim}
\subsection{Notation} Let $\mathbb{R}$ be the set of real numbers, $\mathbb{R}_{\geq 0}:=[0,\infty)$, and $\mathbb{R}_{>0}:=(0,\infty)$. Given two vectors $u\in\mathbb{R}^{n_u}$ and $v\in\mathbb{R}^{n_v}$, we denote the column vector $(u^{\top},v^{\top})^{\top}$ by $(u,v)$ and the inner product $u^{\top}v$ by $\langle u,v \rangle$. For a constant $\rho>0$, we say that a finite set $\bar{\mathcal{A}}\subset\mathbb{R}^n$ is a $\rho$-net of a set $\mathcal{A}\subseteq\mathbb{R}^n$ if for every $a'\in\mathcal{A}$ there exists a $a\in\bar{\mathcal{A}}$ such that $m(a,a)\le \rho$ for a metric $m$ defined in $\mathbb{R}^n$. A continuous function $\alpha:\mathbb{R}\to\mathbb{R}$ is an extended class $\mathcal{K}$ function, denoted by $\alpha \in \mathcal{K}_e$, if it is strictly increasing and $\alpha(0)=0$. 
A continuous function $\beta:\mathbb{R}_{\geq0}\times \mathbb{R}_{\geq 0} \to \mathbb{R}_{\geq 0}$ is a class $\mathcal{KL}$ function, if: (i) $\beta(\cdot,s)$ is a class $\mathcal{K}$ function for each $s\geq 0$; (ii) $\beta(r,\cdot)$ is non-increasing, and (iii) $\beta(r,s)\to 0$ as $s\to \infty$ for each $r\geq 0$.

\subsection{Nonsmooth Control Barrier Functions}

The main objective of this paper is to show that a given set $\mathcal{C}\subseteq\mathbb{R}^{n_x}$ is forward invariant with respect to a system
\begin{equation} \label{eq:sys-inc-gen}
    \dot{x}(t) = f(x(t))    
\end{equation}
which is described by a continuous function $f:\mathbb{R}^{n_x}\to \mathbb{R}^{n_x}$. Continuity of the function $f$ ensures that there exists smooth system trajectories, i.e., there exist solutions $x:[0,T(x(0)))\to \mathbb{R}^{n_x}$ to \eqref{eq:sys-inc-gen} with initial condition $x(0)\in\mathbb{R}^{n_x}$ for a maximal interval of existence $T(x(0))$ \cite[Ch. 3.1]{khalil2002nonlinear}. We later use the notation $\mathcal{I}:=[0,T(x(0)))$ for simplicity.

By forward invariance, we mean that $x(0)\in\mathcal{C}$ implies that $x(t)\in\mathcal{C}$ for all times $t\ge 0$ and for all solutions $x$ to \eqref{eq:sys-inc-gen} with initial condition $x(0)$. To accomplish this, it has to hold that $T(x(0))=\infty$.  We will achieve this via a surrogate function $b:\mathcal{D} \to \mathbb{R}$, where $\mathcal{D}\subseteq \mathbb{R}^{n_x}$ is an open, connected set. This function is known as a candidate control barrier function (CBF), which is used to define the set $\mathcal{C}:=\{x\in\mathcal{D} : b(x) \geq 0  \}$ that we require to be non-empty. Equivalently, the set $\mathcal{C}$ is forward invariant if $x(0)\in\mathcal{C}$ implies that $b(x(t))\geq 0$ for all times $t\in[0,\infty)$ and for all solutions $x$ to \eqref{eq:sys-inc-gen} with initial condition $x(0)$. We call such a function $b$ a CBF for~\eqref{eq:sys-inc-gen}.

In our paper, the candidate CBF we will use is locally Lipschitz continuous, but nonsmooth, which requires nonsmooth analysis tools such as the generalized gradient to handle the non-differentiable points of the nonsmooth function. We recall known results on nonsmooth control barrier functions from \cite{glotfelter2017nonsmooth} which we will use in this paper.

\begin{defn}[Clarke generalized gradient]
    Consider a locally Lipschitz continuous function $b:\mathcal{D}\to \mathbb{R}$. The \textit{Clarke generalized gradient} at $x\in\mathcal{D}$ is the set
    \begin{align} \label{eq:clarke}
        \partial b(x) := \textrm{co}\bigg\{ v\in\mathbb{R}^{n_x} | \exists & x_k \to x,\,  x_k \not\in \mathcal{N}_{b}\nonumber \\ 
        &\textrm{ s.t. } v=\lim_{k\to\infty} \nabla b(x_k) \bigg\},
    \end{align}
    where $\mathcal{N}_b:=\left\{ x\in\mathcal{D} \, | \, \nabla b(x) \textrm{ does not exist} \right\}$ has zero Lebesgue measure, by Rademacher's theorem.
\end{defn}

According to \cite[Thm. 2.5.1]{clarke1990opt}, it was proven that at least one sequence $x_k$ as shown in \eqref{eq:clarke}  exists, which implies that the set $\partial b(x)$ is non-empty for all $x\in\mathcal{D}$. We now recall a result from \cite{glotfelter2017nonsmooth} which we use in this paper.\footnote{The work in \cite{glotfelter2017nonsmooth} focuses on the general case of nonsmooth CBFs $b$ and discontinuous systems $f$. We are here interested in the setting where  $b$ is nonsmooth, but $f$ is continuous so that the results from \cite{glotfelter2017nonsmooth} still apply.}

\begin{prop}[Prop. 2 of \cite{glotfelter2017nonsmooth}] \label{prop:min-cbf}
    Let $\mathcal{P}\subset \mathbb{R}^{n_p}$ be a  finite set and $h:\mathcal{D}\times \mathcal{P} \to \mathbb{R}$ be a locally Lipschitz continuous function that defines a candidate CBF on $\mathcal{D}$.\footnote{The set $\mathcal{P}$ can be thought of as an index set, e.g., to capture a finite set of functions $h(x,1)$, $h(x,2)$, $\hdots$ in which case $\mathcal{P}\subset \mathbb{N}$.} Consider then the minimum $b^{\min}_{\mathcal{P}}(x):= \underset{p\in\mathcal{P}}{\min} \, h(x,p)$ and define the set of active functions $I(x):=\{ p' \in \mathcal{P} \, | \, h(x,p') = \underset{p\in\mathcal{P}}{\min} \, h(x,p) \}$ together with the set of active generalized gradients $\mathcal{E}_{b}(x) := \bigcup_{p\in I(x)} \partial h(x,p)$. If $b^{\min}_{\mathcal{P}}(x)$ is a candidate CBF and there exists $\bar{\alpha}\in \mathcal{K}_e$ such that for every $x \in \mathcal{D}$ and $\xi \in \mathcal{E}_b(x)$, $\langle \xi , f(x) \rangle \geq -\bar{\alpha}(b^{\min}_{\mathcal{P}}(x))$, then $b^{\min}_{\mathcal{P}}(x)$ is a CBF for \eqref{eq:sys-inc-gen}. 
\end{prop}


\section{Motivation and Problem Formulation} \label{sec:mot-prob}
\subsection{Motivation}
We are motivated by automated trajectory tracking and obstacle avoidance of complex shaped robots. Suppose a nominal trajectory tracking  controller has been designed for the system to track a specified state trajectory $x_d:\mathbb{R}_{\ge 0} \to \mathbb{R}^{n_x}$. Concurrent to tracking the user-specified trajectory $x_d$, the robot also has to avoid obstacles. This requires appropriate geometrical representation of the whole robot's body as well as the obstacle(s). 

\begin{figure}[h!]
    \centering
    \includegraphics[width=\linewidth]{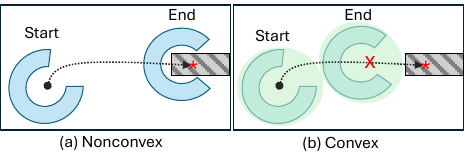}
    \caption{Nonconvex geometric representation (blue) of the C-shaped robot is needed over the spherical representation (green) to move towards the target position {\color{red}$\star$} while avoiding the rectangular obstacle (grey box). The convex representation terminates prematurely at {\color{red} $\times$}. }
    \label{fig:robot-cartoon}
\end{figure}

To this end, let $\mathbb{V}:\mathcal{X}\subseteq \mathbb{R}^{n_x}\to 2^{\mathbb{R}^{n_x}}$  be a set-valued map from the state space $\mathcal{X}$ to the $n_x$ dimensional physical space of the robot; and the set of obstacles be $\mathbb{O}:=\bigcup_{i\in\{1,\dots,N_o\}} \mathbb{O}_i$, where each obstacle is represented by at least one set $\mathbb{O}_i \subseteq \mathbb{R}^{n_x}$. In other words, $\mathbb{V}(x)$ and $\mathbb{O}$ represents the $n_x$ dimensional physical space occupied by the robot at state $x$ and the obstacle, respectively. The formulation in our paper allows for nonconvex $\mathbb{V}(x)$ and $\mathbb{O}$ for close proximity obstacle avoidance, which cannot be achieved by convex geometric representations. We illustrate a use case which motivated this study in Figure \ref{fig:robot-cartoon}.

For collision avoidance, we need to actuate the robot trajectory $x(t)$ such that $\mathbb{V}(x(t)) \bigcap \mathbb{O} = \emptyset$, which requires a notion of the \textit{distance between two sets}, the robot $\mathbb{V}(x)$ and the obstacle(s) $\mathbb{O}$ given by
\begin{equation} \label{eq:distance}
    d(\mathbb{V}(x),\mathbb{O}):= \underset{\substack{v\in \mathbb{V}(x), \\ o\in\mathbb{O}}}{\inf} \| v - o \|,
\end{equation}
where $\|.\|$ denotes the Euclidean distance. Hence, obstacle avoidance is achieved when $d(\mathbb{V}(x),\mathbb{O})>0$. 

\subsection{Problem formulation} \label{sec:prob}
Motivated by the obstacle avoidance problem for a robot with a complex body as described previously, we consider the following system
\begin{equation}\label{eq:system}
    \dot{x} = f(x) + g(x)u,
\end{equation}
where the state is $x\in\mathbb{R}^{n_x}$, the control input is $u\in\mathbb{R}^{n_u}$, and the functions $f:\mathbb{R}^{n_x} \to \mathbb{R}^{n_x}$ and $g:\mathbb{R}^{n_x} \to \mathbb{R}^{n_x\times n_u}$ are continuous. We assume that $g(x)$ has full row rank for each state $x\in\mathbb{R}^{n_x}$ so that $g(x)g(x)^T$ is invertible. This assumption implies that we can fully actuate our robot, e.g., as is the case for omnidirectional robots in our case study. Similar assumptions are often made in the CBF literature \cite{lindemann2020barrier}, but can easily be relaxed for higher-order systems \cite{tan2021high} or unicycle dynamics \cite{lindemann2020control}. We emphasize that we here focus on studying a novel obstacle avoidance technique that uses
efficient sampling-based distance functions. In this paper, we thus aim to solve the following problem. \\
\textbf{Problem (O)}: Given a user-specified trajectory $x_d:\mathbb{R}_{\geq 0} \to \mathbb{R}^{n_x}$, the goal is to design a feedback control law $u^*:\mathbb{R}^{n_x} \to \mathbb{R}^{n_u}$ such that the solution of the closed-loop system $\dot{x}=f(x)+g(x)u^*(x)$ tracks the trajectory $x_d$ in obstacle-free regions and  avoids collision between the robot's body and the obstacles, i.e., $d(\mathbb{V}(x(t)),\mathbb{O})>0$ for all $t\in\mathbb{R}_{\geq 0}$.

\section{Sampling-Based Distance Functions for Obstacle Avoidance via Non-Smooth CBFs } \label{sec:sampling-ncbf}

We achieve automated obstacle avoidance using CBFs, which is a nonlinear control technique to enforce forward invariance of a pre-defined set. However, standard CBF techniques typically use surrogate functions to define this pre-defined set instead of the actual distance function $d(\mathbb{V}(x),\mathbb{O})$. This is for two main reasons: (1) the distance function is difficult to compute even in the case when $\mathbb{V}(x)$ and $\mathbb{O}$ are convex sets, and (2) it is challenging to check whether or not $d(\mathbb{V}(x),\mathbb{O})$ is a valid CBF for the system in equation \eqref{eq:system}. The authors in \cite{thirugnanam2023nonsmooth} consider convex sets $\mathbb{V}(x)$ and $\mathbb{O}$ and propose to re-write the distance function as an optimization problem that subsequently encodes the CBF. However, this proves to be computationally demanding, while checking validity of the CBF remains challenging. Instead, we propose a sampling-based approach to approximate the distance function $d(\mathbb{V}(x),\mathbb{O})$. Such an approach has the benefit that it is easier to analyze, computationally more efficient, and able to deal with nonconvex sets $\mathbb{V}(x)$ and~$\mathbb{O}$.

\subsection{Sampling-Based  Approximation of the Distance Function} \label{sec:sampling_distance}

Assume that we have (densely) sampled the sets $\mathbb{V}(x)$ and $\mathbb{O}$ to obtain a  finite set of  samples $\bar{\mathbb{V}}(x)\subseteq\mathbb{V}(x)$ and $\bar{\mathbb{O}}\subseteq\mathbb{O}$ such that the sampled squared\footnote{We use the squared distance function to obtain differentiability of the individual terms $\| v - o \|^2$ within $d(\bar{\mathbb{V}}(x),\bar{\mathbb{O}})^2$.}  distance function
\begin{align*}
d(\bar{\mathbb{V}}(x),\bar{\mathbb{O}})^2:=\underset{\substack{v\in \bar{\mathbb{V}}(x),  o\in\bar{\mathbb{O}}}}{\min} \| v - o \|^2
\end{align*}
under-approximates the squared distance function $d(\mathbb{V}(x),\mathbb{O})^2$ with a sampling error   $\epsilon\ge 0$  such that
\begin{align}\label{eq:under}
d(\bar{\mathbb{V}}(x),\bar{\mathbb{O}})^2-\epsilon \le d({\mathbb{V}}(x),{\mathbb{O}})^2 
\end{align}
for all $x\in\mathcal{X}$. We note that the sampling error $\epsilon$ is  an implicit function of the shape of the sets $\mathbb{V}(x)$ and $\mathbb{O}$ as well as the sampling routine. In principle, we can use any sampling routine to obtain the sets $\bar{\mathbb{V}}(x)$ and $\bar{\mathbb{O}}$. However, sampling routines that result in small sampling errors with less samples are desirable in practice.  One general approach to construct $\bar{\mathbb{V}}(x)$ and $\bar{\mathbb{O}}$ that satisfy the under-approximation property in equation \eqref{eq:under} is to  construct $\bar{\mathbb{V}}(x)$ and $\bar{\mathbb{O}}$ in a way that densely covers $\mathbb{V}(x)$ and $\mathbb{O}$ using the notion of a  $\rho$-net. 
\begin{lem}\label{lem:1}
 Let $\bar{\mathbb{V}}(x)$ and $\bar{\mathbb{O}}$ be $\rho$-nets of $\mathbb{V}(x)$ and $\mathbb{O}$ for the choices of $\rho:=\epsilon/2$ and $m(a,a'):=\|a-a'\|^2$. Then, the under-approximation property in equation \eqref{eq:under} is satisfied. 

 \begin{proof}
Note first that $\|v-o\|^2=\|v-v'+v'-o+o'-o'\|^2\le \|v'-o'\|^2 + \epsilon$ for each $v$, $v'$, $o$, and $o'$ that are such that $\|v-v'\|^2\le \rho$ and $\|o-o'\|^2\le \rho$  where we used the triangle inequality and the fact that $\rho:=\epsilon/2$.  Next, let $v'$ and $o'$ correspond to the minimum of $d(\mathbb{V}(x),\mathbb{O})$, i.e., be such that $\|v'-o'\|^2=d(\mathbb{V}(x),\mathbb{O})^2$. We know that for each $v'\in \mathbb{V}(x)$ and $o'\in \mathbb{O}$ there exist points $v\in\bar{\mathbb{V}}(x)$ and $o\in \bar{\mathbb{O}}$ such that $\|v-v'\|^2\le \rho$ and $\|o-o'\|^2\le \rho$. Therefore, we know that there exist points $v\in\bar{\mathbb{V}}(x)$ and $o\in \bar{\mathbb{O}}$ such that $\|v-o\|^2 \le d(\mathbb{V}(x),\mathbb{O})^2 + \epsilon$ from which we can conclude that $d(\bar{\mathbb{V}}(x),\bar{\mathbb{O}})^2 \le d(\mathbb{V}(x),\mathbb{O})^2 + \epsilon$ which was to be shown.
 \end{proof}
\end{lem}

To construct $\rho$-nets of $\mathbb{V}(x)$ and $\mathbb{O}$, there are various computational methods to sample $\mathbb{V}(x)$ and $\mathbb{O}$, e.g., gridding or uniform sampling from these sets. The minimum number of samples to construct a $\rho$-nets is given by the covering number  \cite[Chapter 4.2]{vershynin2018high}. Generally, the covering number reveals a relationship between the dimension of the problem (in this case $\mathbb{R}^{n_x}$) and the minimum number of samples required, i.e., for Euclidean balls and many other sets the covering number scales exponentially in its dimension. 


It is obvious that samples from the interiors of $\mathbb{V}(x)$ and $\mathbb{O}$ are not necessarily needed to achieve the guarantee in \eqref{eq:under} as long as there is a $\rho$-net that covers the boundaries of the sets $\mathbb{V}(x)$ and $\mathbb{O}$. In practice, a more efficient method is hence to sample from the boundaries of $\mathbb{V}(x)$ and $\mathbb{O}$ and use the inter-sample distance together with the shape of $\mathbb{V}(x)$ and $\mathbb{O}$  to compute $\epsilon$. While this method may require less samples, we need a sampling routine that can sample from the boundary of a given set. Lastly, we note that computing $d(\bar{\mathbb{V}}(x),\bar{\mathbb{O}})^2$ is computationally tractable and hence a good CBF candidate.

\subsection{Sampling-Based Non-Smooth Control Barrier Functions}

To simplify the following analysis, we write the set $\bar{\mathbb{V}}(x)$ as the Minkowsi sum of a set $\bar{\mathbb{E}}$ that is independent of $x$ and the vector $x$, i.e., such that $\bar{\mathbb{V}}(x):=x \oplus \bar{\mathbb{E}}$.\footnote{We note that such a decomposition is always possible by a unique set $\bar{\mathbb{E}}$ defined as $\bar{\mathbb{E}}:=\bar{\mathbb{V}}(x)\oplus (-x)$.} 
Then, we define the candidate control barrier function
\begin{align}\label{eq:barrier}
b(x,\bar{\mathbb{E}},\bar{\mathbb{O}})
    &:=\underset{\substack{e\in \bar{\mathbb{E}},  o\in\bar{\mathbb{O}}}}{\min} \| x+e- o \|^2-\epsilon-\gamma
\end{align}
where $\gamma\ge 0$ is a user-defined safety margin and $\epsilon \geq 0$ is the sampling-induced error discussed in Section \ref{sec:sampling_distance}. For convenience, we  define $r:=\epsilon+\gamma$. We then denote by 
\begin{align} \label{eq:safeset}
    \mathcal{C}:=\{x\in\mathbb{R}^{n_x}|b(x,\bar{\mathbb{E}},\bar{\mathbb{O}})\ge 0\}
\end{align}
 as our safe set that we aim to render forward invariant. We  note that $b(x,\bar{\mathbb{E}},\bar{\mathbb{O}})=d(\bar{\mathbb{V}}(x),\bar{\mathbb{O}})^2-\epsilon-\gamma$ by which
\begin{align}\label{eq:cons}
b(x,\bar{\mathbb{E}},\bar{\mathbb{O}})\ge 0 \;\;\; \implies \;\;\; d({\mathbb{V}}(x),{\mathbb{O}})\ge \sqrt{\gamma}
\end{align}
using Lemma \ref{lem:1}.
In other words, ensuring non-negativity of the sampling-based candidate control barrier function  ensures that the distance of the robot ${\mathbb{V}}(x)$ to the obstacles ${\mathbb{O}}$ is guaranteed to be greater than the  margin  $\sqrt{\gamma}$.

We  want to guarantee that $b(x,\bar{\mathbb{E}},\bar{\mathbb{O}})$ is also a valid control barrier function. It is immediately clear that $b(x,\bar{\mathbb{E}},\bar{\mathbb{O}})$ is not always differentiable. To analyze this, let us define the set
\begin{align} \label{eq:active-set}
    I(x):=\big\{(e,o)& \in \bar{\mathbb{E}}  \times \bar{\mathbb{O}}| \nonumber \\
    &\| x+e- o \|^2=\underset{\substack{e\in \bar{\mathbb{E}},  o\in\bar{\mathbb{O}}}}{\min} \| x+e- o \|^2\big\},
\end{align}
i.e., the set of active points $(e,o)\in \bar{\mathbb{E}} \times \bar{\mathbb{O}}$ where $\| x+e- o \|^2$ obtains its minimum for a fixed $x$. We can easily conclude that $b(x,\bar{\mathbb{E}},\bar{\mathbb{O}})$ is non-differentiable in $x$ when $|I(x)|>1$. With this in mind, we are now ready to show that $b(x,\bar{\mathbb{E}},\bar{\mathbb{O}})$ is  a valid non-smooth control barrier function.

\begin{thm}\label{thm:1}
    Consider the system \eqref{eq:system}, the candidate control barrier function $b(x,\bar{\mathbb{E}},\bar{\mathbb{O}})$ defined in \eqref{eq:barrier}, and the state domain $\mathcal{D}:=\{x\in \mathbb{R}^{n_x}\,| \,  b(x,\bar{\mathbb{E}},\bar{\mathbb{O}})\in (-\bar{r},\infty)\}$ where $0<\bar{r}<r$. Furthermore, let the set $\mathcal{C}$ defined in \eqref{eq:safeset} be compact. Then, for each state $x\in\mathcal{D}$, it holds that 
    \begin{align}\label{eq:CBF_cond}
        \sup_{u\in \mathbb{R}^{n_u}}\inf_{\zeta\in\mathcal{E}(x)} \langle \zeta, f(x)+g(x)u\rangle\ge -\alpha(b(x,\bar{\mathbb{E}},\bar{\mathbb{O}}))
    \end{align}
    where $\alpha\in\mathcal{K}_{e}$ and 
    \begin{align}\label{eq:E_set_CBF}
    \mathcal{E}(x):=\bigcup_{(e,o)\in I(x)} \, 2( x+e- o).
\end{align} 
Furthermore, if $x(0)\in\mathcal{C}$ and if $u:\mathcal{D}\to\mathbb{R}^{n_u}$ is a continuous function that, for all $x\in\mathcal{D}$, satisfies
    \begin{align}\label{eq:CBF_cond_u}
        u(x)\in\big\{u\in\mathbb{R}^{n_u}\,\,\inf_{\zeta\in\mathcal{E}(x)} \langle \zeta, f(x)&+g(x)u\rangle\nonumber\\
        &-\alpha(b(x,\bar{\mathbb{E}},\bar{\mathbb{O}})\big\},
    \end{align}
    then it holds that $d({\mathbb{V}}(x(t)),{\mathbb{O}})\ge \sqrt{\gamma}$ for all $t\ge 0$ and for all solutions $x(t)$ of the system \eqref{eq:system} under $u$.


    \begin{proof} 
    \emph{Step 1. } First recall that $x\in\mathcal{D}$ so that  $b(x,\bar{\mathbb{E}},\bar{\mathbb{O}})=\underset{\substack{e\in \bar{\mathbb{E}},  o\in\bar{\mathbb{O}}}}{\min} \| x+e- o \|^2-r\in (-\bar{r},\infty)$ by definition. From here, it follows that $\underset{\substack{e\in \bar{\mathbb{E}},  o\in\bar{\mathbb{O}}}}{\min} \| x+e- o \|^2\in (0,\infty)$ since $r-\bar{r}>0$. Consequently, for all $(e,o)\in I(x)$, we conclude that the vector $2( x+e- o)$ cannot be zero so that the set $\mathcal{E}(x)$ in equation \eqref{eq:E_set_CBF} cannot contain zero. Additionally, we recall that the function $g(x)g(x)^T$ is invertible by assumption. From these  observations, we can see that condition \eqref{eq:CBF_cond} holds for all $x\in\mathcal{D}$ and for any extended class $\mathcal{K}$ function $\alpha$.

    \emph{Step 2. } Now, let $x:\mathcal{I}\mapsto \mathcal{D}$ be a solution to the initial value problem of \eqref{eq:system} with initial condition $x_0\in \mathcal{C}$ and where $\mathcal{I}\subseteq \mathbb{R}_{\ge 0}$ is the maximum domain of  $x$.  We next show that $x(t)\in\mathcal{C}$ for all $t\in \mathcal{I}$ if $x(0)\in\mathcal{C}$ and if $u:\mathcal{D}\to\mathbb{R}^{n_u}$ is a continuous function that  satisfies \eqref{eq:CBF_cond_u} by applying Proposition \ref{prop:min-cbf}. Specifically, since we are dealing with a non-smooth candidate control barrier function, we need to consider the generalized gradient of $b(x,\bar{\mathbb{E}},\bar{\mathbb{O}})$. Therefore, we define the set
\begin{align*}
    \mathcal{E}(x)&:= \bigcup_{(e,o)\in I(x)} \partial\, ( \| x+e- o \|^2-r )
    =2(x+e- o)
\end{align*}
as the union of the generalized gradients of the functions $\| x+e- o \|^2-r$ for the active points $(e,0)\in I(x)$. We  note that $u(x)$ always exists (as shown in Step 1) so that we can  apply Proposition \ref{prop:min-cbf} which gives us the desired result. What remains to be shown is that the solution $x$ is  that $\mathcal{I}=\mathbb{R}_{\ge 0}$.  In that regard, recall that $\mathcal{C}$ is assumed to be a compact set. By  \cite[Theorem 3.3]{khalil2002nonlinear}, it then follows that $\mathcal{I}=\mathbb{R}_{\ge 0}$.

       \emph{Step 3. }  From Lemma \ref{lem:1} and the observation in equation \eqref{eq:cons} it finally follows that $d({\mathbb{V}}(x(t)),{\mathbb{O}})\ge \sqrt{\gamma}$.
    \end{proof}
\end{thm}

We note that Theorem \ref{thm:1} guarantees forward invariance for compact sets $\mathcal{C}$. Compactness of $\mathcal{C}$ is solely needed  to ensure that robot trajectories $x(t)$ exist for all times $t\ge 0$, i.e., that solutions $x$ to the ordinary differential equation in \eqref{eq:system} are complete. If $\mathcal{C}$ is not compact, but the control law $u$ ensures completeness of solutions $x$ in another way, then Theorem \ref{thm:1} still remains valid. In our setting, we can easily accomplish compact sets $\mathcal{C}$ by considering a `virtual' obstacle that surrounds the workspace of the robot.

\subsection{CBF-based control law}
The existence of a control barrier function generates a family of controllers given by the set in \eqref{eq:CBF_cond_u}. Further, Theorem \ref{thm:1} also showed that if the control law $u$ is continuous, forward invariance of the set $\mathcal{C}$ is guaranteed and hence obstacle avoidance is achieved, i.e., $d(\mathbb{V}(x(t)),\mathbb{O})>0$ for $t\geq 0$. 

To solve the collision-free trajectory tracking problem formulated as \textbf{Problem (O)} in Section \ref{sec:prob}, we consider the use of the CBF-generated control laws in the context of a safety filter. Given a nominal control law $u_{d}:\mathbb{R}^{n_x}\to \mathbb{R}^{n_u}$ which has been designed to track a desired trajectory with no obstacle avoidance capabilities, we find a minimally invasive control law that satisfies the CBF conditions, by leveraging the benefits of the following quadratic program (QP) 
\begin{equation} \label{eq:QP-CBF-control}
    u^*(x) \in \underset{u\in\mathbb{R}^{n_u}}{\arg\min} \|u-u_{d}(x)\|^2 \; \textrm{s.t. }   \eqref{eq:CBF_cond} \textrm{ holds}.
\end{equation}

The QP is convex and provides a computationally lightweight algorithm for fast online implementation. This formulation was motivated by the prevalence of QP-based CBF controllers for Lipschitz systems with differentiable CBFs \cite{ames2016control}, where closed-form control laws can be derived. In our case where we have nonsmooth CBFs, ensuring the regularity of the control law $u^*(x)$ such that the set $\mathcal{C}$ is forward invariant with respect to the closed loop system is challenging. Nonetheless, we guarantee that we have a collision-free trajectory tracking controller under the assumptions stated in the theorem below. 

\begin{thm}
    Consider the system \eqref{eq:system}, the candidate control barrier function $b(x,\bar{\mathbb{E}},\bar{\mathbb{O}})$ defined in \eqref{eq:barrier}, the state domain $\mathcal{D}:=\{x\in \mathbb{R}^{n_x}\,| \,  b(x,\bar{\mathbb{E}},\bar{\mathbb{O}})\in (-\bar{r},\infty)\}$ where $0<\bar{r}<r$. Furthermore, let the set $\mathcal{C}$ defined in \eqref{eq:safeset} be compact. Suppose \begin{enumerate}[(i)]
        \item there is a nominal control law $u_d:\mathcal{D}\to \mathbb{R}^{n_u}$ such that the solution to $\dot{x}=f(x)+g(x)u_d(x)$ satisfies \begin{equation} \label{eq:tracking-error-0}
        \|x(t)-x_d(t)\|\leq \beta(\|x(0)-x_d(0)\|,t)
    \end{equation}
    for all $t\ge 0$ and  $(x(0),x_d(0))\in\mathbb{R}^{2n_x}$ where $\beta\in\mathcal{KL}$;
    \item the system is initially safe, i.e., $x(0)\in\mathcal{C}$; and
    \item there exists a continuous  minimizer $u^*:\mathcal{D}\to\mathbb{R}^{n_u}$ for the QP in \eqref{eq:QP-CBF-control} for some $\alpha\in\mathcal{K}_{e}$.
    \end{enumerate}
    Then, \textbf{Problem (O)} is solved. 

    \begin{proof}
        Let the continuous function $u^*:\mathcal{D}\to\mathbb{R}^{n_x}$ be the minimizer which solves the QP in \eqref{eq:QP-CBF-control} and $x(0)\in\mathcal{C}$. First, when there are no obstacles, i.e., $x\in\mathcal{C}$, then we have from (i) that $u_d(x)$ satisfies \eqref{eq:CBF_cond_u} and is the minimizer of QP \eqref{eq:QP-CBF-control}, i.e., $u^*(x)=u_d(x)$. Further, we also have from (i) that the desired trajectory $x_d$ is tracked according to \eqref{eq:tracking-error-0}.  
        Next, since a continuous $u^*(x)$ is the minimizer which solves the QP in \eqref{eq:QP-CBF-control}, the inequality \eqref{eq:CBF_cond} must hold with a continuous $u:\mathcal{D}\to \mathbb{R}^{n_u}$ satisfying \eqref{eq:CBF_cond_u}. Then, by using Theorem \ref{thm:1}, we have that $d({\mathbb{V}}(x(t)),{\mathbb{O}})\ge \sqrt{\gamma}$. Hence, we have shown that $u^*(x)$ solves \textbf{Problem (O)}.
    \end{proof}
\end{thm}

We  assumed that the minimizer $u^*:\mathcal{D}\to\mathbb{R}^{n_u}$ for QP \eqref{eq:QP-CBF-control} is continuous such that forward invariance of the set $\mathcal{C}$ can be guaranteed by Theorem \ref{thm:1}. In general, the minimizer $u$ is not guaranteed to be continuous even with the assumed continuity of the system dynamics. Nonetheless, as we see in our simulation study in Section \ref{sec:sim}, we solve \textbf{Problem (O)}. In fact, continuity of $u^*$ is often assumed in the CBF literature \cite{ames2016control}.   Further investigations will have to be made on the continuity of $u^*$ and the closed-loop system dynamics. Another idea is to modify the QP \eqref{eq:QP-CBF-control} to ensure continuity. Works such as \cite{alyaseen2024safety, isaly2024feasibility} are good starting points in ensuring the feasibility of the modified QP and the continuity of the minimizer $u^*(x)$, which we leave as future work.

\section{Incorporating  Disturbances} \label{sec:disturbance-errors}

So far, we assumed that the system dynamics in equation \eqref{eq:system} were perfectly known. In this section, we extend our sampling-based non-smooth CBF framework to deal with structured and unstructured disturbances on the  dynamics.

To model disturbances, consider now the perturbed system
\begin{equation}\label{eq:system_}
    \dot{x} = f(x) + g(x)u+d(x,t),
\end{equation}
where $d:\mathbb{R}^{n_x}\times \mathbb{R}_{\ge 0}\to \mathbb{R}^{n_x}$ is unknown. We consider the cases where $d$ has no and where $d$ has some structure. 

Let us start with the case where $d$ has no structure. Here, the function $d$ is assumed to be completely unknown except for knowledge of an upper bound for $d$, as is standard in the robust control literature \cite{freeman2008robust}. Indeed, assume that we know an  upper bound $D:\mathbb{R}^{n_x}\times\mathbb{R}_{\ge 0} \to \mathbb{R}_{\ge 0}$ such that $\|d(x,t)\|\le D(x,t)$ for all $(x,t)\in\mathbb{R}^{n_x}\times \mathbb{R}_{\ge 0}$. As shown in the next result, we can now simply tighten the condition \eqref{eq:CBF_cond} to obtain the same safety guarantees as before.

\begin{thm}\label{thm:1_}
    Consider the system \eqref{eq:system_} with $\|d(x,t)\|\le D(x,t)$ for all $(x,t)\in\mathbb{R}^{n_x}\times \mathbb{R}_{\ge 0}$, the candidate control barrier function $b(x,\bar{\mathbb{E}},\bar{\mathbb{O}})$ defined in \eqref{eq:barrier}, and the state domain $\mathcal{D}:=\{x\in \mathbb{R}^{n_x}\,| \,  b(x,\bar{\mathbb{E}},\bar{\mathbb{O}})\in (-\bar{r},\infty)\}$ where $0<\bar{r}<r$. Furthermore, let the set $\mathcal{C}$ defined in \eqref{eq:safeset} be compact. Then, for each state $x\in\mathcal{D}$, it holds that 
 \begin{align}\label{eq:CBF_cond_d}
        \sup_{u\in \mathbb{R}^{n_u}}\inf_{\zeta\in\mathcal{E}(x)} \langle \zeta, f(x)+g(x)u\rangle -  \| &\zeta\| D(x,t) \nonumber \\
        &\ge -\alpha(b(x,\bar{\mathbb{E}},\bar{\mathbb{O}})),
\end{align}
    where $\alpha\in\mathcal{K}_e$  and $\mathcal{E}(x)$ is defined in \eqref{eq:E_set_CBF}. Furthermore, if $x(0)\in\mathcal{C}$ and if $u:\mathcal{D}\to\mathbb{R}^{n_u}$ is a continuous function that, for all $x\in\mathcal{D}$, satisfies
    \begin{align}\label{eq:CBF_cond_u_d}
    \begin{split}
        u(x)\in \big\{u\in\mathbb{R}^{n_u} \left| \right. &\inf_{\zeta\in\mathcal{E}(x)} \langle \zeta, f(x)+g(x)u\rangle\\&- \| \zeta\|D(x,t)\ge -\alpha(b(x,\bar{\mathbb{E}},\bar{\mathbb{O}})\big\},
    \end{split}
    \end{align}
 then it holds that $d({\mathbb{V}}(x(t)),{\mathbb{O}})\ge \sqrt{\gamma}$ for all $t\ge 0$ and for all solutions $x(t)$ of the system \eqref{eq:system} under $u$.
    
    \begin{proof} 
    The proof follows almost the same steps as the proof of Theorem \ref{thm:1}. However, here the  satisfaction of  \eqref{eq:CBF_cond_d} (instead of  \eqref{eq:CBF_cond} in Theorem \ref{thm:1})   implies that
\begin{align*}
        \sup_{u\in \mathbb{R}^{n_u}}\inf_{\zeta\in\mathcal{E}(x)} \langle \zeta, f(x)+g(x)u+d(x,t)\rangle\ge -\alpha(b(x,\bar{\mathbb{E}},\bar{\mathbb{O}}))
\end{align*}
for any $d(x,t)$ that is such that $\|d(x,t)\|\le D(x,t)$ since $-\zeta d(x,t)\le \|\zeta\|D(x,t)$. The rest of the proof is the same as Theorem \ref{thm:1} except for that we can now consider the system in \eqref{eq:system_} (instead of the system in \eqref{eq:system} in Theorem \ref{thm:1}).  
    \end{proof}
\end{thm}

Let us now consider the case where $d$ has some structure in the sense that $d(x,t):=g(x)e(x)$ where  $e(x):=u_\text{app}(x)-u(x)$ is a tracking error between the applied and the desired control input. For instance, in many applications the safety controller $u(x)$ is forwarded to a low-level PID controller that is such that $u_\text{app}(x)\approx u(x)$. We assume here that we know the tracking error, i.e., that we have knowledge of an upper bound $E:\mathbb{R}^{n_x}\times\mathbb{R}_{\ge 0}$ such that $\|e(x,t)\|\le E(x,t)$ for all $(x,t)\in\mathbb{R}^{n_x}\times\mathbb{R}_{\ge 0}$. We obtain a similar result as before by tightening the condition  \eqref{eq:CBF_cond}.

\begin{thm}\label{thm:1__}
    Consider the system \eqref{eq:system_} where $d(x,t):=g(x)e(x)$ and $e(x):=u_\text{app}(x)-u(x)$ with $\|e(x,t)\|\le E(x,t)$ for all $(x,t)\in\mathbb{R}^{n_x}\times\mathbb{R}_{\ge 0}$, the candidate control barrier function $b(x,\bar{\mathbb{E}},\bar{\mathbb{O}})$ defined in \eqref{eq:barrier}, and the state domain $\mathcal{D}:=\{x\in \mathbb{R}^{n_x}\,| \,  b(x,\bar{\mathbb{E}},\bar{\mathbb{O}})\in (-\bar{r},\infty)\}$ where $0<\bar{r}<r$. Furthermore, let the set $\mathcal{C}$ defined in \eqref{eq:safeset} be compact. Then, for each state $x\in\mathcal{D}$, it holds that 
\begin{align}\label{eq:CBF_cond_dd}
        \sup_{u\in \mathbb{R}^{n_u}}\inf_{\zeta\in\mathcal{E}(x)} \langle \zeta, f(x)+g(x)u\rangle - \| & \zeta g(x)\|E(x,t) \nonumber \\
        &\ge -\alpha(b(x,\bar{\mathbb{E}},\bar{\mathbb{O}})),
\end{align}
    where $\alpha\in\mathcal{K}_e$ and $\mathcal{E}(x)$ is defined in \eqref{eq:E_set_CBF}. Furthermore, if $x(0)\in\mathcal{C}$ and if $u:\mathcal{D}\to\mathbb{R}^{n_u}$ is a continuous function that, for all $x\in\mathcal{D}$, satisfies
    \begin{align}\label{eq:CBF_cond_u_dd}
    \begin{split}
        u(x)\in \big\{u\in\mathbb{R}^{n_u} \left| \right. &\inf_{\zeta\in\mathcal{E}(x)} \langle \zeta, f(x)+g(x)u\rangle\\& \hspace{-0.5cm} - \| \zeta g(x)\|E(x,t)\ge -\alpha(b(x,\bar{\mathbb{E}},\bar{\mathbb{O}})\big\},
    \end{split}
    \end{align}
    then it holds that $d({\mathbb{V}}(x(t)),{\mathbb{O}})\ge \sqrt{\gamma}$ for all $t\ge 0$ and for all solutions $x(t)$ of the system \eqref{eq:system} under $u$.
    
    \begin{proof} 
    The proof follows almost the same steps as the proof of Theorem \ref{thm:1_} and is omitted for brevity.
    \end{proof}
\end{thm}

\section{Simulation study} \label{sec:sim}
\begin{figure*}
\centering
\includegraphics[width=0.9\linewidth]{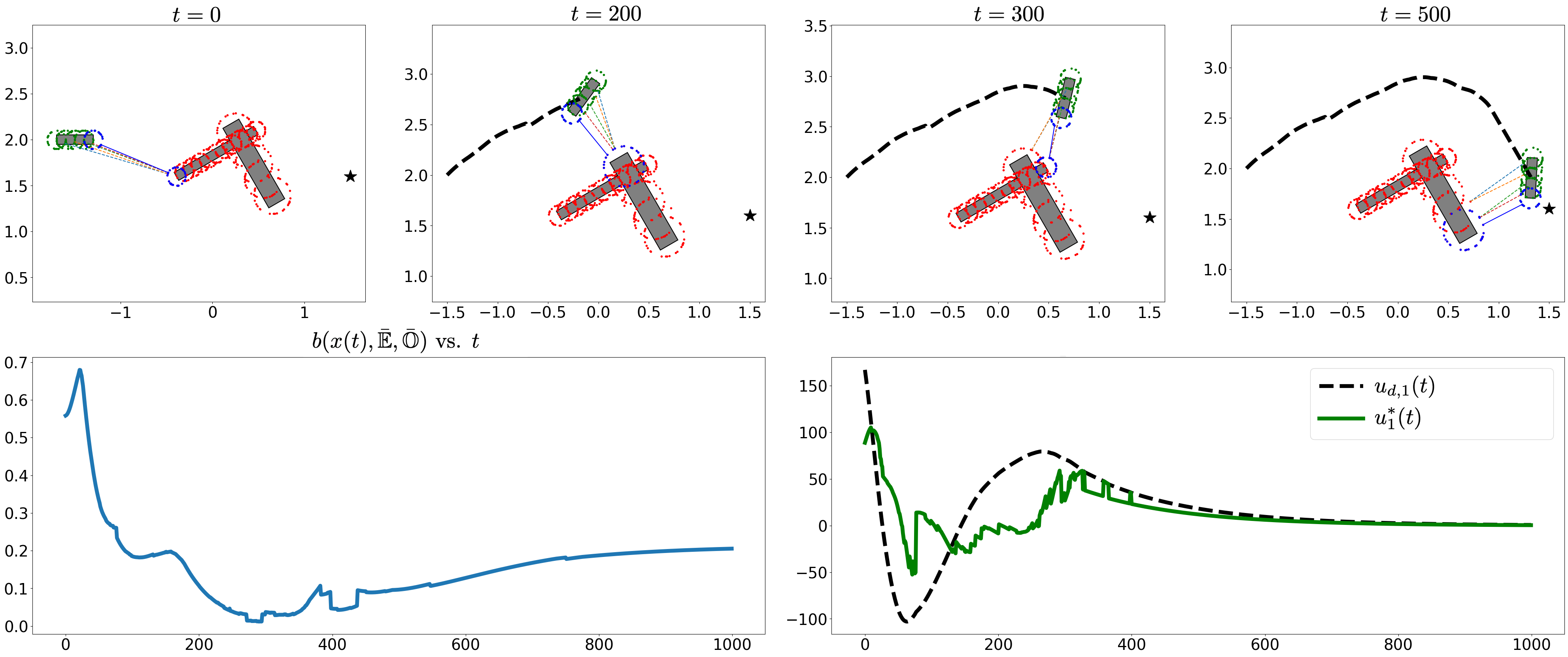}
\caption{(Top): Snapshots of robot trajectory at times $t\in \{0,200,300,500\}$s \textbf{without disturbances}. {(Bottom left): CBF $b(x(t),\bar{\mathbb{E}},\bar{\mathbb{O}})$ vs. time $t$.} (Bottom right):  Nominal control law $u_d$ and QP-CBF-based control law $u^*$ from \eqref{eq:QP-CBF-control}.}
\label{fig_1}
\end{figure*}

We consider an omnidirectional three-wheeled robot navigating through a space with nonconvex obstacles, see Figure \ref{fig_1} (top display). The robot position $(x_1,x_2)\in\mathbb{R}^2$ and its orientation $x_3\in\mathbb{R}$ is represented in a global reference frame. The kinematic model is taken from\cite{liu2008omni} and given as
\begin{equation} \label{eq:omni}
    \dot{x} = G(x) (B^T)^{-1}  u, 
\end{equation}
where $G(x):=\left(\begin{array}{ccc} \cos(x_3) & -\sin(x_3) & 0 \\ \sin(x_3) & \cos(x_3) & 0 \\ 0 & 0 & 1 \end{array}\right)$ and $B:=\left(\begin{array}{ccc} 0 & r \cos(\pi/6) & -r \cos(\pi/6) \\ -r & r\sin(\pi/6) & r\sin(\pi/6) \\ lr & lr & lr \end{array}\right)$ describe  a rotation matrix and the robot's geometry, respectively, with $l:=0.2$ and $r:=0.02$ being the radius of the robot's body and the radius of each wheel. Each component $u_i$ of the control input $u\in\mathbb{R}^3$ is the angular velocity of a wheel. By choosing the state to be $x:=(x_1,x_2,x_3)$, the robot kinematic model in \eqref{eq:omni} is in the form of system \eqref{eq:system} with $f(x)=0$ and $g(x)=G(x) (B^T)^{-1}$. We use a PID controller to generate the tracking control law $u_d=(u_{d,1},u_{d,2},u_{d,3})$ that guarantees reaching a goal position from an initial position, but does not by itself guarantee collision avoidance. Throughout, we select a safety margin of $\gamma:=0.05$.

\begin{figure*}
\centering
\includegraphics[width=0.9\linewidth]{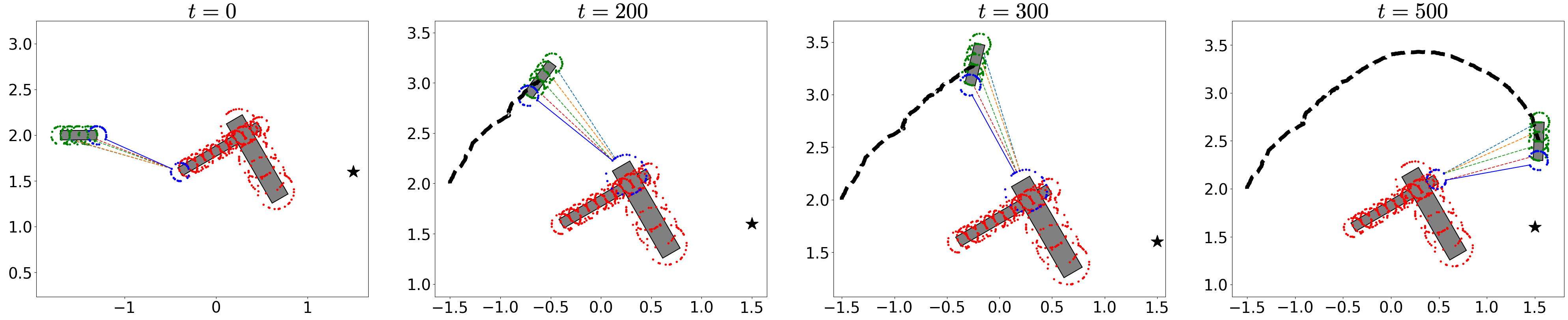}
\caption{Snapshots of robot trajectory at times $t\in \{0,200,300,500\}$s \textbf{with disturbances}. }
\label{fig_2}
\end{figure*}
\begin{figure}[h!]
    \centering
    \includegraphics[width=0.4\linewidth]{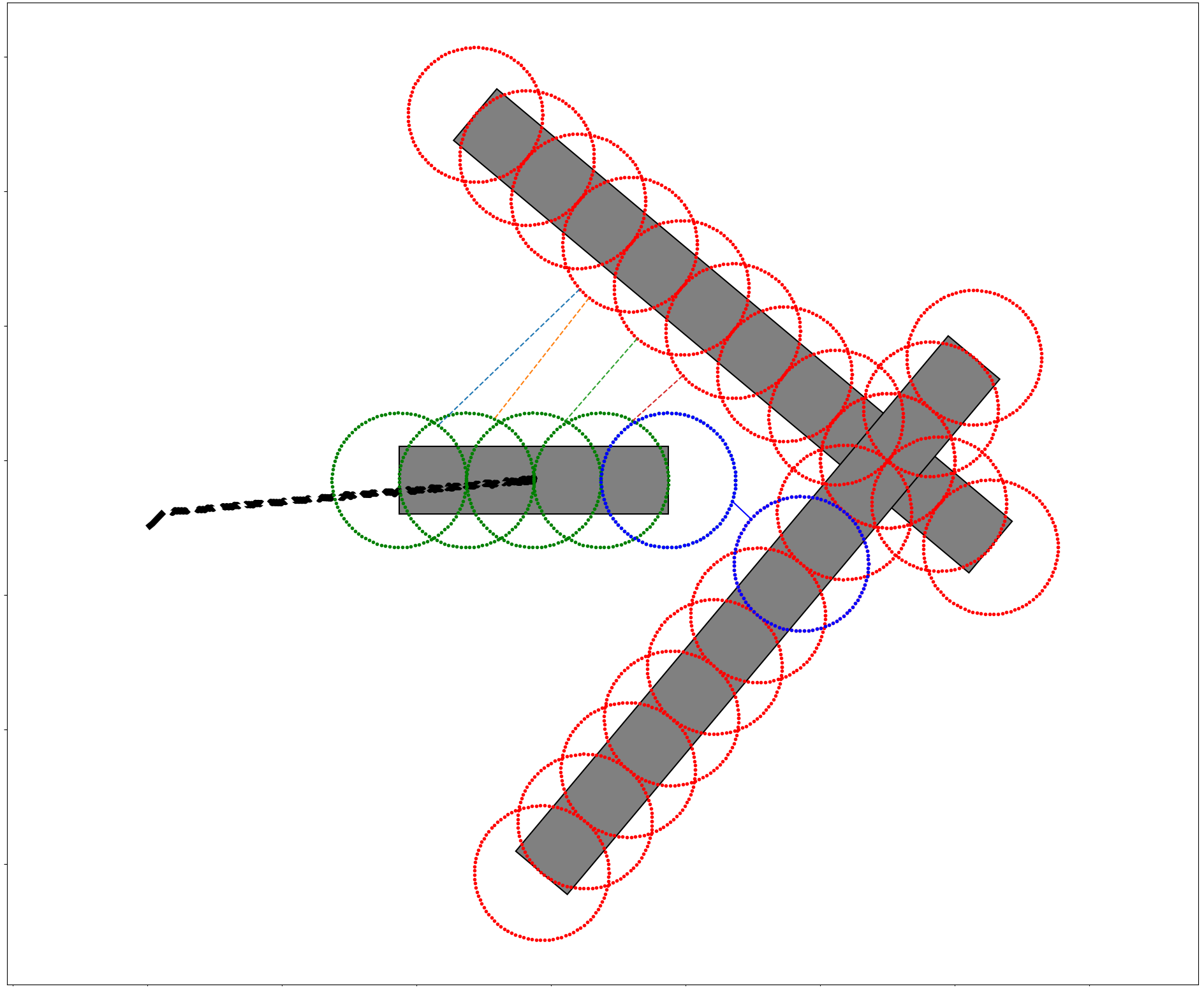}
    \vspace{-1em}
    \caption{Robot in deadlock to study trade-offs, see Figure  \ref{fig_4}.}
    \label{fig_3}
\end{figure}
\begin{figure}[h!]
\centering
\includegraphics[width=\linewidth]{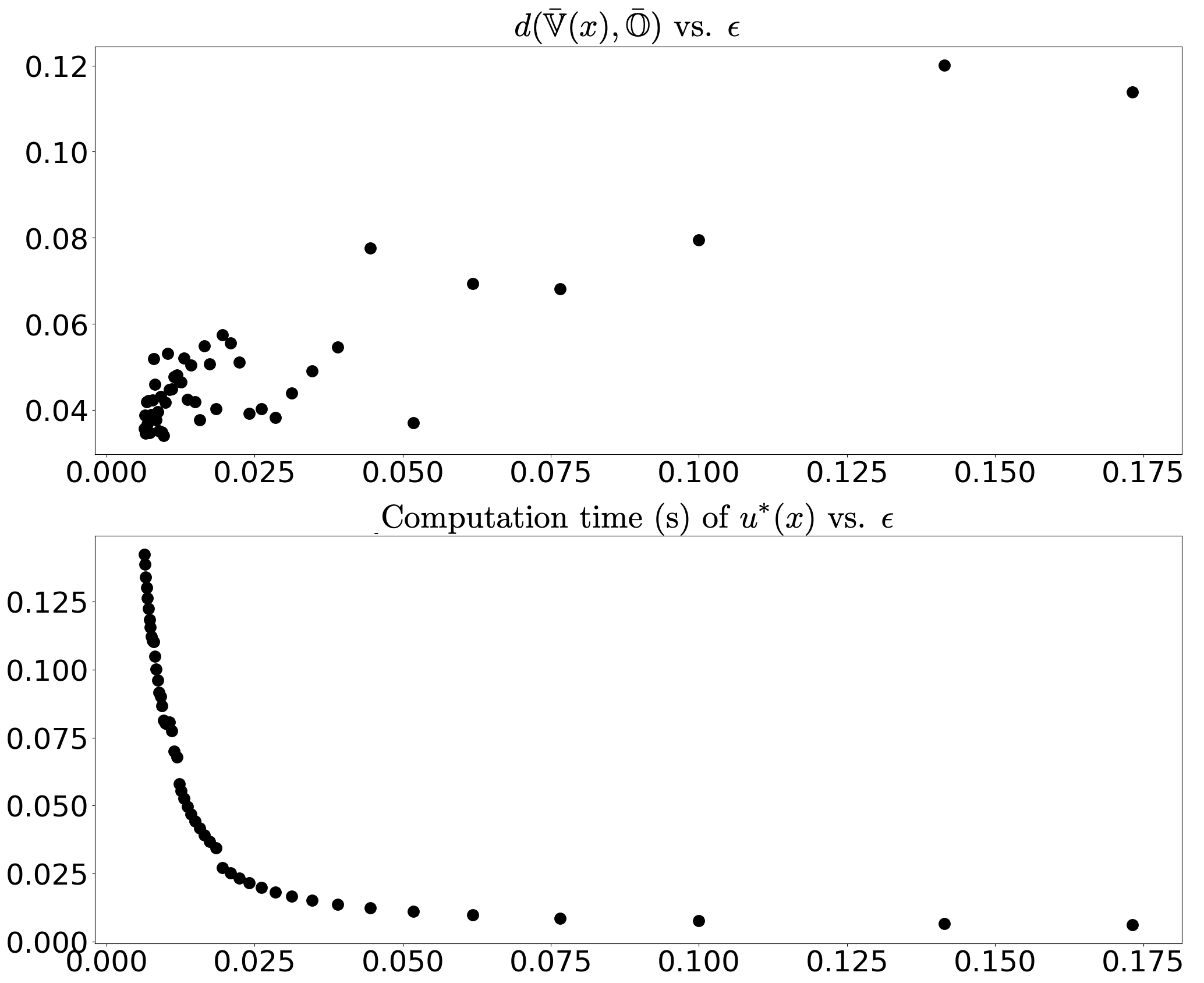}
\caption{ Tradeoff between the sampling error $\epsilon$ with (i) achievable minimum distance between sampled points on robot and obstacles $d(\bar{\mathbb{V}}(x),\bar{\mathbb{O}})$ (Top plot); and (ii) Time taken to compute safe control law $u^*$ (Bottom plot).}
\label{fig_4}
\end{figure}

\textbf{Obstacle Sampling. } We have two overlapping, rectangle-shaped obstacles that together form a nonconvex shape which the robot has to avoid. We construct a union of circles that tightly enclose the shapes of the robot and the obstacles and hence define the sets ${\mathbb{V}}(x)$ and ${\mathbb{O}}$. The green and red dotted circles denote $\bar{\mathbb{V}}(x)$ and $\bar{\mathbb{O}}$, respectively, see Figure \ref{fig_1} (top display).  Specifically, as elaborated already at the end of Section \ref{sec:sampling_distance}, we construct $\rho$-nets $\bar{\mathbb{V}}(x)$ and $\bar{\mathbb{O}}$ that (for computational reasons) only cover the boundaries of the sets $\mathbb{V}(x)$ and $\mathbb{O}$. To accomplish this, we employ two different techniques to obtain the samples in $\bar{\mathbb{V}}(x)$ and $\bar{\mathbb{O}}$. Our first technique randomly samples (with uniform distribution) $N$ samples from the boundaries of ${\mathbb{V}}(x)$ and ${\mathbb{O}}$. Our second technique, instead, creates a uniform grid of the boundaries of ${\mathbb{V}}(x)$ and ${\mathbb{O}}$. For both techniques, we then compute the sampling error $\epsilon$ afterwards. We use the first technique to show our results for  obstacle avoidance with and without disturbances (the next two paragraphs) and the second technique for analyzing trade-offs (last paragraph).

\textbf{Nonconvex Obstacle Avoidance. } We solve the convex optimization problem in \eqref{eq:QP-CBF-control} to obtain the controller $u^*$ that achieves the navigation task. We plot the results for one experiment in Figure \ref{fig_1}.  The top display shows the robot trajectories at times $0$, $200$, $300$, and $500$. We also indicate in blue the closest samples between $\bar{\mathbb{V}}(x)$ and $\bar{\mathbb{O}}$ that define the set $\mathcal{E}(x)$. The bottom left display shows the evolution of $b(x(t),\bar{\mathbb{E}},\bar{\mathbb{O}})$, while the bottom right display shows the evolution of $u^*(x(t))$. It can be seen that obstacle avoidance reaches its goal location. While our controller guarantees obstacle avoidance, it is worth mentioning that the robot may in principle get stuck and not reach its goal position, e.g., when being driven into the region where the  rectangle-shaped obstacles form an obtuse angle. Recovering from such deadlock positions can be achieved by using deadlock recovery strategies and updating the tracking controller $u_d$.

\textbf{Disturbances. } We repeat the same experiment as before, but add unknown random disturbances $d(t)$ to the system dynamics. Here, $d(t)$ is drawn from a uniform distribution with support $[-0.4,0.4]^3$ so that $D:=0.4\cdot \sqrt{3}$. We then solve the convex optimization problem in \eqref{eq:QP-CBF-control} by replacing the constraint \eqref{eq:CBF_cond} with \eqref{eq:CBF_cond_d}. The results are shown in Figure \ref{fig_2}, where a more conservative robot trajectory is visible. 

\textbf{Trade-offs. } Lastly, we perform a set of experiments in which we vary the number of samples in $\bar{\mathbb{V}}(x)$ and $\bar{\mathbb{O}}$  so that we can study trade-offs between  the approximation error $\epsilon$ (which is a function of the number of samples $N$) and (1) the minimum distance to obstacle, and (2) the computation time. To create a situation where we can fairly compare the minimum distance, we create an experiment where the robot is deadlocked, see Figure \ref{fig_3}. The tradeoff curves are shown in Figure \ref{fig_4}. As expected, with increasing sampling error $\epsilon$,  conservatism increases while computation time decreases.

\section{Conclusions and future work} \label{sec:conclude}

In this work, we provided an efficient solution to nonconvex obstacle avoidance by computing the distance between the robot and obstacles (described by nonconvex shapes) via sampling-based distance functions. We then quantified the sampling error and showed that, for certain systems, sampling-based distance functions are valid nonsmooth control barrier functions. We validated our method on an omnidirectional robot that had to navigate among a nonconvex obstacle. Future work will include addressing the regularity properties of the QP-based control law, the incorporation of input constraints, relaxing the assumption of invertible input dynamics $g$, and the integration of state estimation errors.

\bibliographystyle{ieeetr}
\bibliography{CBF_obstacle_avoidance.bib}

\end{document}